\documentclass[lettersize,journal]{IEEEtran}

\usepackage{graphicx}
\usepackage{amsmath}
\usepackage{amssymb}
\usepackage{booktabs}
\usepackage{caption}
\usepackage{xcolor}
\usepackage{diagbox}
\usepackage[export]{adjustbox}
\usepackage[pagebackref,breaklinks,colorlinks]{hyperref}
\usepackage[ruled, linesnumbered]{algorithm2e}
\usepackage{multirow}

\begin{document}

\title{Resource Efficient Neural Networks Using Hessian Based Pruning}

\author{

Jack Chong\textsuperscript{1*}, 
Manas Gupta\textsuperscript{2*},
Lihui Chen\textsuperscript{3}

\vspace{3mm}

~\IEEEmembership{
\small 
\textsuperscript{1} School of Electrical and Electronic Engineering (EEE), Nanyang Technological University (NTU), Singapore, \\
\textsuperscript{2} Institute for Infocomm Research (I2R), Agency for Science, Technology and Research (A*STAR), Singapore, \\
\textsuperscript{3} Centre for Information Sciences and Systems (CISS), School of Electrical and Electronic Engineering (EEE), Nanyang Technological University (NTU), Singapore \\
}

\vspace{3mm}

\small jchong046@e.ntu.edu.sg, manas\_gupta@i2r.a-star.edu.sg, elhchen@ntu.edu.sg

\IEEEcompsocitemizethanks{\IEEEcompsocthanksitem * Equal contribution}
}

\maketitle

\begin{abstract}
  Neural network pruning is a practical way for reducing the size of trained models and the number of floating point operations (FLOPs). One way of pruning is to use the relative Hessian trace to calculate sensitivity of each channel, as compared to the more common magnitude pruning approach. However, the stochastic approach used to estimate the Hessian trace needs to iterate over many times before it can converge. This can be time-consuming when used for larger models with many millions of parameters. To address this problem, we modify the existing approach by estimating the Hessian trace using FP16 precision instead of FP32. We test the modified approach (EHAP) on ResNet-32/ResNet-56/WideResNet-28-8 trained on CIFAR10/CIFAR100 image classification tasks and achieve faster computation of the Hessian trace. Specifically, our modified approach can achieve speed ups ranging from 17\% to as much as 44\% during our experiments on different combinations of model architectures and GPU devices. Our modified approach also takes up $\sim$40\% less GPU memory when pruning ResNet-32 and ResNet-56 models, which allows for a larger Hessian batch size to be used for estimating the Hessian trace. Meanwhile, we also present the results of pruning using both FP16 and FP32 Hessian trace calculation and show that there is no noticeable accuracy differences between the two. Overall, it is a simple and effective way to compute the relative Hessian trace faster without sacrificing on pruned model performance. We also present a full pipeline using EHAP and quantization aware training (QAT), using INT8 QAT to compress the network further after pruning. In particular, we use symmetric quantization for the weights and asymmetric quantization for the activations. 
  The framework has been open-sourced and the code is available at https://github.com/JackkChong/Resource-Efficient-Neural-Networks-Using-Hessian-Based-Pruning.
\end{abstract}

\section{Introduction}
In recent years, neural networks have become more powerful in many tasks such as computer vision and natural language processing~\cite{lecun_bengio_hinton_2015}. This is partly due to larger models with many millions of parameters and increasing input sizes. Some notable examples for image classification on ImageNet are EfficientNet-B7 with 66 million parameters~\cite{EfficientNet2019} and ViT-G/14 vision transformer with 1843 million parameters~\cite{ScalingViTs2021}. However, it has been shown that these models with millions of parameters are usually over-parameterized~\cite{sze2017arXiv}. Moreover, it is very hard to deploy these models on cloud servers or edge devices due to their large size and bigger memory footprint. Since most deep neural networks contain redundant neurons, one way to make these models deploy-able is to compress them through neural network pruning. Since then, much research has been conducted to determine the best way to prune a neural network, such as the magnitude-based approach or the second order Hessian-based approach.

An important challenge to pruning is to find out which parameters are insensitive and can be pruned away without severe degradation of the model's performance. One way of doing so is to use the relative Hessian trace to calculate the sensitivity of each channel, in contrast to the standard magnitude pruning approach. The seminal work of~\cite{zheweiyao2021arXiv} proposed Hessian Aware Pruning, which makes use of the stochastic Hutchinson trace estimator to determine the relative Hessian trace of each channel.

However, the Hutchinson trace estimator used to estimate the Hessian trace needs to iterate hundreds of times before it can converge to a stable value. This can be time-consuming when calculating the Hessian trace for larger models with many billions of parameters. In addition, the compression ratio of the pruned model is non-deterministic and can fluctuate within a few percentage points. The Hessian trace may need to be computed multiple times before pruning the model if a specific compression ratio is desired. To address this problem, we modify the existing approach in HAP by estimating the Hessian trace using FP16 precision instead of FP32. Our main contributions are:
\begin{itemize}
    \item We test the default FP32 approach on 3 different neural network models (ResNet-32/ResNet-56/WideResNet-28-8) and record the average time spent to estimate the Hessian trace.
    \item We test our modified FP16 approach and compare the average time spent to estimate the Hessian trace against the default approach.
    \item We record the peak memory statistics of the GPU when estimating the Hessian trace using both FP32 and FP16 approaches.
    \item We compare the pruning results using both FP32 and FP16 approaches.
    \item We present a full pipeline using EHAP and quantization aware training together. INT8 quantization aware training (QAT) is done to compress the network further after pruning.
\end{itemize}

\section{Related Work}
In recent years, the size of neural networks have rapidly increased and there is a growing need for faster inference on edge devices~\cite{tombrown2020arXiv, 9516010, 8818358, 8756206, Liu2021ARA, 8693518}. Gradually, many approaches to make neural networks more compact and efficient have started to emerge \cite{cheng2017survey, 9478787, almahairi2016dynamic,ashok2017n2n,i2016squeezenet,pham2018efficient}. In this paper, we briefly discuss the related work on pruning and quantization.

\subsection{Pruning}
Pruning can be grouped into two categories: structured pruning and unstructured pruning. Unstructured pruning aims to prune away redundant neurons without any structure. Unfortunately, this can result in sparse matrix operations which are difficult to speed up using GPUs~\cite{blalock2020arXiv}. To address this, structured pruning removes groups of redundant neurons in a structured way so that matrix operations can still be conducted efficiently. However, the difficulty lies in severe accuracy degradation when the model is pruned to a large degree. In both methods, the important challenge is to determine which parameters are insensitive to pruning. Many methods have been proposed in the literature to address this utilising regularization approaches \cite{pmlr-v119-kusupati20a, autobalanced, ContinuousSparsification2020, 9398648, wang2021neural}, first or second order methods \cite{NIPS1989_250,NIPS1992_647, bert}, gradient-based techniques \cite{lee2018snip, Wang2020Picking, acdc}, similarity approaches \cite{Srinivas_2015}, sensitivity and feedback \cite{Lin2020Dynamic, molchanov2016pruning, LIU2020Dynamic, jorge2021progressive, 9097925}, and magnitude based approaches \cite{pmlr-v119-evci20a, lee2021layeradaptive, Zhu2018ToPO, Strom97sparseconnection, Park2020LookaheadAF, gupta2022complexity}. 

One of the most popular ways is magnitude-based pruning. At the heart of this approach is the assumption that small parameters are redundant and can be pruned away. For example,~\cite{zhu2017arXiv} explores the use of magnitude based pruning to achieve a significant reduction in model size and only a marginal drop in accuracy. Another variant was used in~\cite{zhuangliu2017arXiv} to prune redundant neurons using the scaling factor of batch normalization layers. In particular, those channels with smaller scaling factors are seen as unimportant and were pruned away. However, an important drawback of magnitude-based pruning is that parameters with smaller magnitudes can still be quite sensitive to pruning. A second-order Taylor series expansion of the model's loss function shows that the change to the loss depends on not just the magnitude of the weights, but also the second-order Hessian~\cite{LeCun1989OptimalBD}. In particular, parameters with small magnitudes may still possess a high degree of second-order sensitivity.

To account for the second-order component in the Taylor series expansion, several Hessian-based pruning methods have emerged. The Hessian diagonal was used as the sensitivity metric in~\cite{LeCun1989OptimalBD}.~\cite{Hassibi1993OptimalBS} also used the Hessian diagonal, but considered off-diagonal components and proved a correlation with the Hessian inverse. However, one important drawback of these methods was that it results in unstructured pruning, which is difficult to accelerate with GPUs. There are also other second-order pruning methods that do not use the Hessian. For example, EigenDamage~\cite{eigendamage2019arXiv} uses the Gauss-Newton operator instead of the Hessian and the authors estimate the GN operator using Kronecker products.

For this paper, we focus on Hessian Aware Pruning from~\cite{zheweiyao2021arXiv} to conduct second-order structured pruning of a pretrained network. In this method, a stochastic Hutchinson trace estimator is used to determine the relative Hessian trace. The result is then used to calculate the sensitivity of each channel.

\subsection{Quantization}
By default, neural networks are trained using FP32 precision. However, most networks do not need such a high level of precision to infer well. A model's weights and activations can be quantized from floating point into lower numerical precision so that it can boost inference time without a significant drop in accuracy~\cite{tailinliang2021arXiv}. Currently, the two main focus areas for quantization are Post-Training Quantization (PTQ) and Quantization Aware Training (QAT). The difference between the two is whether the parameters in the quantized model are fine-tuned after quantization.

In PTQ, the model is quantized directly without any fine-tuning~\cite{zhang2022arXiv}. As such, PTQ methods usually do not rely heavily on additional training data. However, it comes at the cost of lower accuracy because quantization can degrade the accuracy of the model significantly without fine-tuning.

In QAT, fine-tuning is done after quantization by adding quantizer blocks into the pretrained model. These quantizer blocks simulate the quantization noise by conducting quantize and de-quantize operations~\cite{markusnagel2021arXiv}. Although QAT requires additional time and computational resources to fine-tune the model, it results in higher accuracy of the model.

For this paper, we use QAT using the framework in~\cite{hawq3} to quantize the pruned models using INT8 precision and report our results in Section \ref{exp}.

\begin{figure}[h]
\centering
\includegraphics[width=0.7\linewidth]{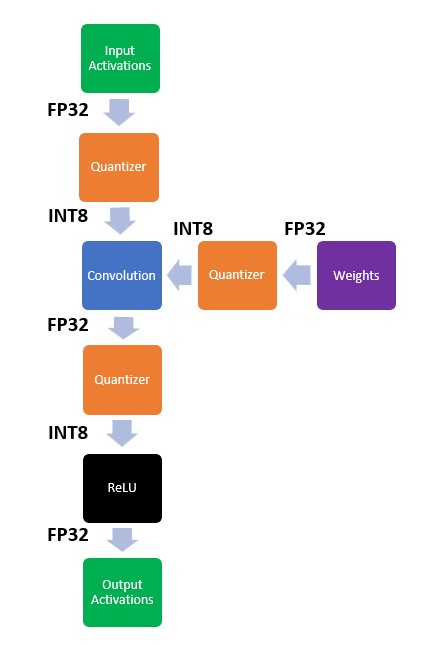}
\caption{An illustration of simulated quantization based forward pass in floating point precision. The quantizer blocks conduct quantize and de-quantize operations to induce quantization errors into the computational graph so that the model can optimize on them during QAT.}
\label{fig:quantization2} 
\end{figure}

\section{Methodology}
The focus of this paper is on supervised learning tasks, where we aim to minimize the empirical risk by solving the following optimization problem:
\begin{equation}
\label{eq:empiricalrisk}
    L(w) = \frac{1}{N} \sum_{i=1}^{N} l(x_i,y_i,w),
\end{equation}
where \(w~\epsilon~\mathbb{R}^{N}\) refers to trainable model parameters, \(l(x_i,y_i,w)\) refers to the loss for each input sample \(x_i\), \(y_i\) refers to the target label and \(N\) refers to the number of training samples. Before pruning, we make the assumption that the model has been pretrained and converged to a local minima. This means that the gradient, \(\nabla_{w}L(w) = 0\) and the Hessian is Positive Semi-Definite (PSD). The main objective is to prune as many parameters as possible to compress the model without significant accuracy degradation.

\begin{algorithm}[t]
\caption{Hutchinson's Method}\label{alg:Hutchinson}
    \DontPrintSemicolon 
    \KwIn{Parameters $\theta$, number of iterations $n_{v}$}
    \KwOut{Hessian trace approximation $\mathbf{E}[v^{T}Hv]$}
    Compute the gradient of $\theta$ by backpropagation, $g_{i} = \frac{dL}{dW_{i}}$\;
    \For{\(i=1,2,\hdots n_{v}\)}
    {
        Draw a random vector $v$ from the Rademacher distribution\;
        Compute $Hv$ by backpropagation, $Hv=\frac{d(gv)}{d\theta}$\;
        Compute $v^{T}Hv$ by taking dot product between $v$ and $Hv$\;
    }
    Compute $\mathbf{E}[v^{T}Hv]$ by taking mean of all $v^{T}Hv$ over $n_{v}$ iterations\;
\end{algorithm}

Let \(\Delta w~\epsilon~\mathbb{R}^{N}\) represent the pruning perturbation such that the pruned weights go to zero. The corresponding change in loss from Taylor Series expansion is represented as:
\begin{equation}
\label{eq:taylorseries}
    \Delta L = L(w+ \Delta w) - L(w) = g^{T}\Delta w + \frac{1}{2}\Delta w^{T}H\Delta w + O(||\Delta w||^{3})
\end{equation}
where \(g\) represents the gradient of the loss function \(L\) w.r.t. weights \(w\) and \(H\) represents the second-order derivative (Hessian). We let \(g=0\) and the Hessian be PSD for a pretrained network that has converged to a local minimum. We also assume higher order terms can be ignored~\cite{Hassibi1993OptimalBS}.

To determine the weights that give minimum perturbation to the loss, we present the following optimization problem:
\begin{equation}
\label{eq:lagrangian}
\underset{\Delta w}{min}\frac{1}{2}\Delta w^{T}H\Delta w = \frac{1}{2}
\begin{bmatrix}
    \Delta w_{p} \\
    \Delta w_{l}
\end{bmatrix}^{T}
\begin{bmatrix}
    H_{p,p} & H_{p,l} \\
    H_{l,p} & H_{l,l}
\end{bmatrix}
\begin{bmatrix}
    \Delta w_{p} \\
    \Delta w_{l}
\end{bmatrix}
\end{equation}

where \(\Delta w_{p}\) and \(\Delta w_{l}\) denote the perturbation to the weights of the pruned 
channels (p-channels) and un-pruned channels (l-channels) respectively, \(H_{l,p}\) represents
cross Hessian w.r.t. l-channels and p-channels, \(H_{p,p}\) represents Hessian w.r.t. p-channels 
only and \(H_{l,l}\) represents Hessian w.r.t. l-channels only.
Since we are pruning p-channels, we subject our optimization problem to the following constraint:
\begin{equation}
\label{eq:lagrangian_constraint}
    \Delta w_{p} + w_{p} = 0
\end{equation}

By using the Lagrangian method to solve the constrained optimization problem, we get the following (refer to Appendix for details):
\begin{equation}
    \frac{1}{2}\Delta w^{T}H\Delta w = \frac{1}{2}w_{p}^{T}(H_{p,p}-H_{p,l}H_{l,l}^{-1}H_{l,p})w_{p}
\end{equation}
which gives us the change to the model loss when a group of parameters are pruned. Next, we dive into how Hessian Aware Pruning is conducted in~\cite{zheweiyao2021arXiv}.

\subsection{Hessian Aware Pruning (HAP)}
According to~\cite{zheweiyao2021arXiv}, we use the following approximation for the perturbation to the model loss (i.e. the sensitivity metric):
\begin{equation}
    \frac{1}{2}\Delta w^{T}H\Delta w
    \approx \frac{1}{2}w_{p}^{T}\frac{Trace(H_{p,p})}{p}w_{p} = \frac{Trace(H_{p,p})}{2p}||w_{p}||_{2}^{2}
\end{equation}
where \(Trace(H_{p,p})\) refers to the trace of the Hessian block for p-channels. We can approximate this efficiently using a randomized numerical method known as Hutchinson's Method~\cite{Bai1996SomeLM} (Algorithm \ref{alg:Hutchinson}).

\begin{algorithm}[t]
\caption{Efficient Hutchinson's Method}\label{alg:Hutchinson2}
    \DontPrintSemicolon 
    \KwIn{Parameters $\theta$, number of iterations $n_{v}$}
    \KwOut{Scaled Hessian trace approximation $\mathbf{E}[sv^{T}Hv]$}
    Initialize gradient scaler and default scaling factor $s=2^{16}$\;
    Compute the scaled gradient of $\theta$ by FP16 backpropagation, $sg_{i} = s\frac{dL}{dW_{i}}$\;
    Unscale the gradient, $g_{i} = \frac{sg_{i}}{s}$\;
    Reduce the scaling factor to $s = 2^{8}$\;
    \For{\(i=1,2,\hdots n_{v}\)}
    {
        Draw a random vector $v$ from the Rademacher distribution\;
        Compute scaled $Hv$ by FP16 backpropagation, $sHv=s\frac{d(gv)}{d\theta}$\;
        Compute $sv^{T}Hv$ by taking dot product between $v$ and $sHv$\;
    }
    Compute $\mathbf{E}[sv^{T}Hv]$ by taking mean of all $sv^{T}Hv$ over $n_{v}$ iterations\;
\end{algorithm}

By using Hutchinson's Method, we can approximate the Hessian trace without computing the full Hessian. In particular, we can show that (refer to Appendix for details):
\begin{equation}
\label{eq:Hutchinson_Ex}
    Tr(H) = \mathbf{E}[v^{T}Hv]
\end{equation}
and therefore, we use this identity to calculate the second-order sensitivity metric for each channel:
\begin{equation}
    \frac{Trace(H_{p,p})}{2p}||w_{p}||_{2}^{2} = \frac{\mathbf{E}[v^{T}Hv]}{2p}||w_{p}||_{2}^{2}
\end{equation}

\begin{table*}[h]
\centering
\caption{HAP and EHAP Results for CIFAR-10. There are no noticeable accuracy differences between the two approaches.}
\label{CIFAR10_EHAP_Results}
\begin{tabular}[t]{r|ccc}
\toprule
\textbf{CIFAR-10} & \textbf{Original Acc.} & \textbf{Pruned Acc. (HAP)} & \textbf{Pruned Acc. (EHAP)} \\
\midrule
ResNet-32 & 93.83\% $\pm$ 0.09\% & 88.47\% $\pm$ 0.25\% & 88.89\% $\pm$ 0.23\% \\
ResNet-56 & 94.32\% $\pm$ 0.21\% & 90.47\% $\pm$ 0.26\% & 89.83\% $\pm$ 0.42\% \\
WideResNet-28-8 & 96.23\% $\pm$ 0.09\% & 95.14\% $\pm$ 0.08\% & 95.22\% $\pm$ 0.09\% \\
\bottomrule
\end{tabular}
\end{table*}

\begin{table*}[h]
\centering
\caption{HAP and EHAP Results for CIFAR-100. There are no noticeable accuracy differences between the two approaches.}
\label{CIFAR100_EHAP_Results}
\begin{tabular}[t]{r|ccc}
\toprule
\textbf{CIFAR-100} & \textbf{Original Acc.} & \textbf{Pruned Acc. (HAP)} & \textbf{Pruned Acc. (EHAP)} \\
\midrule
ResNet-32 & 71.00\% $\pm$ 0.44\% & 55.33\% $\pm$ 0.58\% & 55.01\% $\pm$ 0.44\% \\
ResNet-56 & 72.89\% $\pm$ 0.19\% & 60.87\% $\pm$ 0.32\% & 60.99\% $\pm$ 0.18\% \\
WideResNet-28-8 & 81.01\% $\pm$ 0.24\% & 73.63\% $\pm$ 0.97\% & 73.67\% $\pm$ 0.32\% \\
\bottomrule
\end{tabular}
\end{table*}

\begin{table*}[h]
\centering
\caption{Timing benchmarks for Hutchinson's Method using an NVIDIA GeForce RTX2060 device. We show here that EHAP can compute the relative Hessian trace faster in all cases as compared to HAP.}
\label{Timings_1}
\begin{tabular}[t]{r|rrc}
\toprule
\textbf{CIFAR-100} & \textbf{Avg. Time} & \textbf{\% Speed Up} & \textbf{Memory Used} \\
\midrule
ResNet-32 (HAP) & 86.970 s & \multirow{2}{4em}{18.40\%} & 2.402 GB \\
ResNet-32 (EHAP) & 70.968 s & & 1.407 GB \\
ResNet-56 (HAP) & 146.659 s & \multirow{2}{4em}{17.80\%} & 3.859 GB \\
ResNet-56 (EHAP) & 120.557 s & & 2.124 GB \\
WRN-28-8 (HAP) & 277.476 s & \multirow{2}{4em}{32.36\%} & 4.073 GB \\
WRN-28-8 (EHAP) & 187.684 s & & 3.678 GB \\
\bottomrule
\end{tabular}
\end{table*}

\subsection{Efficient Hessian Aware Pruning (EHAP)}
The authors in~\cite{zheweiyao2021arXiv} used default FP32 precision to compute the Hessian trace with Hutchinson's Method. In their experiments with ResNet50 on ImageNet, the longest time taken to approximate the Hessian trace was three minutes. However, the compression ratio cannot be directly controlled and may fluctuate within a few percentage points during pruning. If a specific compression ratio is desired, Hutchinson's Method must be run several times. This can be time-consuming if we were to scale up a model to several hundred million parameters and want to run Hutchinson's Method multiple times.

To address this problem, we propose to run the backpropagation operations within Hutchinson's Method in FP16 precision. By leveraging PyTorch's Automatic Mixed Precision (AMP) package~\cite{PyTorchamp}, we can approximate the Hessian trace faster using the modified approach (Algorithm \ref{alg:Hutchinson2}). In more detail, we leverage AMP's autocasting context manager to enable backpropagation to run in FP16 during Hutchinson's Method. We also adopt AMP's gradient scaler to scale gradients during the backpropagation process. The rationale behind scaling is to avoid underflow or overflow. Without scaling, values which are too small will flush to zero (underflow) and those which are too big will turn into NaNs (overflow)~\cite{PyTorchamp}. After tuning, we find that the optimal scaling factor to use for second-order backpropagation in our experiments is $2^{8}$.

In addition, we do not unscale the second-order gradients because the scale does not affect our approximation. This is because we only require the relative Hessian trace to compare between different channels' sensitivities, so we are not concerned with the absolute value of the Hessian trace approximation. Overall, we find that using FP16 precision is a simple way to approximate the Hessian trace faster.

\subsection{Quantization Aware Training (QAT)}
We follow the framework provided in~\cite{hawq3} to conduct INT8 quantization for our pruned models. Our goal is to conduct INT8 QAT for our models after pruning. Below, we explain uniform quantization and the addition of quantizer blocks to insert quantize and de-quantize operations into the computational graph.

\begin{table*}[ht]
\centering
\caption{Timing benchmarks for Hutchinson's Method using an NVIDIA GeForce RTX3090 device. We show here that EHAP can compute the relative Hessian trace faster in all cases as compared to HAP.}
\label{Timings_2}
\begin{tabular}[t]{r|rrc}
\toprule
\textbf{CIFAR-100} & \textbf{Avg. Time} & \textbf{\% Speed Up} & \textbf{Memory Used} \\
\midrule
ResNet-32 (HAP) & 34.722 s & \multirow{2}{4em}{41.36\%} & 2.401 GB \\
ResNet-32 (EHAP) & 29.770 s & & 1.408 GB \\
ResNet-56 (HAP) & 63.231 s & \multirow{2}{4em}{44.57\%} & 3.837 GB \\
ResNet-56 (EHAP) & 46.537 s & & 2.127 GB \\
WRN-28-8 (HAP) & 163.785 s & \multirow{2}{4em}{25.79\%} & 4.956 GB \\
WRN-28-8 (EHAP) & 112.575 s & & 3.678 GB \\
\bottomrule
\end{tabular}
\end{table*}

\begin{table*}[ht]
\centering
\caption{QAT Results for CIFAR-10. We show here that the accuracies achieved after QAT are comparable to those after pruning.}
\label{CIFAR10_QAT_Results}
\begin{tabular}[t]{r|ccc}
\toprule
\textbf{CIFAR-10} & \textbf{Original Acc.} & \textbf{Pruned Acc.} & \textbf{Quantized Acc.} \\
\midrule
ResNet-32 & 93.83\% $\pm$ 0.09\% & 88.47\% $\pm$ 0.25\% & 88.52\% $\pm$ 0.27\% \\
ResNet-56 & 94.32\% $\pm$ 0.21\% & 90.47\% $\pm$ 0.26\% & 89.48\% $\pm$ 1.27\% \\
WideResNet-28-8 & 96.23\% $\pm$ 0.09\% & 95.14\% $\pm$ 0.08\% & 94.87\% $\pm$ 0.18\% \\
\bottomrule
\end{tabular}
\end{table*}

\begin{table*}[ht]
\centering
\caption{QAT Results for CIFAR-100. We show here that, for this case as well, the accuracies achieved after QAT are comparable to those after pruning.}
\label{CIFAR100_QAT_Results}
\begin{tabular}[t]{r|ccc}
\toprule
\textbf{CIFAR-100} & \textbf{Original Acc.} & \textbf{Pruned Acc.} & \textbf{Quantized Acc.} \\
\midrule
ResNet-32 & 71.00\% $\pm$ 0.44\% & 55.33\% $\pm$ 0.58\% & 54.96\% $\pm$ 0.54\% \\
ResNet-56 & 72.89\% $\pm$ 0.19\% & 60.87\% $\pm$ 0.32\% & 60.88\% $\pm$ 0.26\% \\
WideResNet-28-8 & 81.01\% $\pm$ 0.24\% & 73.63\% $\pm$ 0.97\% & 73.72\% $\pm$ 0.81\% \\
\bottomrule
\end{tabular}
\end{table*}

\subsection{Uniform Quantization}
We present the uniform quantization formula as shown below:
\begin{equation}
\label{eq:quantize}
    Q(r) = Int(r/S) - Z
\end{equation}
where $Q$ refers to the quantization function, $r$ is the FP32 weight / activation that we want to quantize, $S$ is a real-valued scaling factor and $Z$ is the quantized value that will represent real-value zero (i.e. zero point). The $Int$ operation is simply a rounding operation to nearest integer. Since the quantized values are uniformly spaced, this method of quantization is called uniform quantization.

An important part of uniform quantization is calibration, where we select the clipping range $[\alpha,\beta]$ so that we can determine the appropriate scaling factor $S$. The easiest way to choose the clipping range is to select the maximum and minimum values of the signal, where $\alpha=r_{min}$ and $\beta=r_{max}$. This corresponds to asymmetric quantization since the clipping range is not symmetric to the origin. An alternative way to choose the clipping range is to let $-\alpha=\beta=max(|r_{max}|,|r_{min}|)$, which corresponds to symmetric quantization. 

In this paper, we adopt symmetric quantization for weights. However, we use asymmetric quantization for activations after ReLU because those activations will always be non-negative. Using asymmetric quantization will maximise the entire 8-bitwidth since we only need to represent positive values for activations.

To recover the floating point values from the quantized values in Equation \ref{eq:quantize}, we have the de-quantization formula shown below:
\begin{equation}
\label{eq:de-quantize}
    r \approx S(Q(r) + Z)
\end{equation}
Note that the de-quantize operation cannot exactly recover the floating point value due to rounding operation in the previous quantize operation. Next we describe how we apply Equations \ref{eq:quantize} and \ref{eq:de-quantize} to add quantizer blocks and conduct QAT. 

\subsection{Adding Quantizer Blocks}
When running experiments, it is easier to conduct QAT on general-purpose hardware instead of the actual quantized device. We do so by using quantizer blocks to insert quantization effects into the computational graph~\cite{hawq3}. During QAT, the quantization errors will then appear in the loss and be accounted for during gradient descent. A schematic diagram of this is shown in \autoref{fig:quantization2}. In more detail, we conduct QAT by simulating quantization on our general-purpose floating point hardware. For both weights and activations, we simply need to pass them through the quantizer blocks during forward pass to induce quantization effects into the computational graph.

\section{Experiments}
\label{exp}

For all our experiments with ResNet-32 / ResNet-56 / WideResNet-28-8, we target a compression ratio of 10\% i.e. sparsifying the model parameters by 90\%. The experimental settings, resulting file sizes, FLOPs and compression ratios can be found in the Appendix. 

\subsection{EHAP Results for CIFAR-10 and CIFAR-100}
We test EHAP on CIFAR-10 \cite{CIFAR-10} and CIFAR-100 \cite{CIFAR-10} using ResNet-32, ResNet-56 and WideResNet-28-8 and show the results in \autoref{CIFAR10_EHAP_Results} and \autoref{CIFAR100_EHAP_Results}. In particular, we report the original accuracy for each model type and compare between the pruned accuracies using EHAP and HAP.

From \autoref{CIFAR10_EHAP_Results} and \autoref{CIFAR100_EHAP_Results}, we show that EHAP (the modified approach) achieves equivalent accuracy results when compared to using HAP (the standard approach). In some cases, EHAP even achieves higher accuracy than HAP on both CIFAR-10 and CIFAR-100 datasets, despite using half the precision for Hessian trace calculation. This outcome happens because layers are pruned based on the relative Hessian trace only. Thus, we do not need a high level of precision for Hutchinson's Method to achieve good results.

We also compare the average time taken to run Hutchinson's Method using both the standard approach and the modified approach in \autoref{Timings_1} and \autoref{Timings_2}.  In particular, we report the average time taken to compute the relative Hessian trace, the percentage speed up and the max memory allocated by the GPU device for Hutchinson's Method using EHAP and HAP. From \autoref{Timings_1} and \autoref{Timings_2}, we achieve speed ups (of upto 44.5\%) in all of our model architectures and on both types of GPU devices. The speed ups can be seen on both the older RTX 2060 device and the newer RTX 3090 device. In addition, we also save on the amount of GPU memory used (of upto 1.7 GB), which can be helpful for GPU devices with limited memory.

\subsection{QAT Results for CIFAR-10 and CIFAR-100}
We conduct QAT for ResNet-32, ResNet-56 and WideResNet-28-8 and show the results below in \autoref{CIFAR10_QAT_Results} and \autoref{CIFAR100_QAT_Results}. In particular, we report the original, pruned and quantized accuracies in the tables for easy comparison. The quantized accuracies shown are the accuracies that we can expect from the model when we deploy to the actual quantized device. It can be seen that QAT does not have an adverse impact on the model accuracy after pruning, achieving very similar accuracy results. In some cases, the QAT accuracy is actually slightly higher than the pruned accuracy. We also achieve smaller file sizes after QAT since we only save the quantized weights and biases and their corresponding scaling factors.

\section{Limitations and Future Work}
We have showcased an integrated pipeline for pruning and quantization utilizing more efficient Hessian aware pruning. A limitation in this work is that the gradient scale used for second-order FP16 backpropagation in EHAP is static and was tuned manually by hand. However, dynamic gradient scaling could be used so that we do not have to tune the scale manually. 

Another area of future work is to modify the quantization scheme to be more sophisticated by using Hessian based quantization such that the bitwidth of each layer is not constant and varies as per the Hessian rule.

\section{Conclusion}

The existing method used to compute the relative Hessian trace in default FP32 can be time-consuming for larger models if we need to run the pruning script multiple times. We propose a modified approach that computes the relative Hessian trace using FP16 precision. From the results obtained using CIFAR-10 and CIFAR-100 benchmark data sets in Section \ref{exp}, we show that our modified approach is more efficient and consistently faster (upto 44.5\%) than the standard approach used to approximate the relative Hessian trace. We also do not compromise on pruned model accuracy using our modified approach at the same compression ratio. In addition, we save on a modest amount of GPU memory used (upto 1.7 GB) during our modified approach, which enables a larger Hessian batch size to be used to improve the accuracy of our Hessian trace approximation. Overall, it is a simple and effective way to compute the relative Hessian trace faster without sacrificing on pruned model accuracy.

\small
\bibliographystyle{IEEEtran}
\bibliography{References}

\newpage
\begin{minipage}{\textwidth}
    \parbox[c][20cm][c]{\linewidth}
    {
    
\section{Appendix}

\subsection{Solving Equation \ref{eq:lagrangian}}
\label{cha:Lagrangian}
We begin by forming the Lagrangian equation and determining its saddle points, similar to~\cite{zheweiyao2021arXiv}. \\
\begin{equation}
    L = \frac{1}{2}\Delta w^{T}H\Delta w + \lambda^{T}(\Delta w_{p} + w_{p})
\end{equation}

\begin{equation}
    \frac{\delta L}{\delta \Delta w} = H\Delta w + \begin{bmatrix} \lambda \\ 0 \end{bmatrix} = 0
\end{equation}

\begin{equation}
\label{eq:appendix_1}
    \begin{bmatrix}
        H_{p,p} & H_{p,l} \\
        H_{l,p} & H_{l,l}
    \end{bmatrix}
    \begin{bmatrix}
        \Delta w_{p} \\
        \Delta w_{l}
    \end{bmatrix}
    +
    \begin{bmatrix}
        \lambda \\
        0
    \end{bmatrix}
    = 0
\end{equation} \\

where \(\lambda~\epsilon~\mathbb{R}^{p}\) refers to the Lagrange multipliers for each pruned parameter. After expanding the above Equation \ref{eq:appendix_1}, we can get Equations \ref{eq:appendix_2} and \ref{eq:appendix_3} below:

\begin{equation}
\label{eq:appendix_2}
    H_{p,p}\Delta w_{p} + H_{p,l}\Delta w_{l} + \lambda = 0
\end{equation}

\begin{equation}
\label{eq:appendix_3}
    H_{l,p}\Delta w_{p} + H_{l,l}\Delta w_{l} = 0
\end{equation} \\

We then apply the constraint in Equation \ref{eq:lagrangian_constraint} to Equation \ref{eq:appendix_3}.

\begin{equation}
    -H_{l,p}w_{p} + H_{l,l}\Delta w_{l} = 0
\end{equation}

\begin{equation}
\label{eq:appendix_4}
    \Delta w_{l} = H_{l,l}^{-1}H_{l,p}w_{p}
\end{equation} \\

Equation \ref{eq:appendix_4} tells us what the optimal change to the unpruned parameters ($w_l$) should be if we prune a set of weights ($w_p$). We insert Equation \ref{eq:appendix_4} into Equation \ref{eq:lagrangian} and obtain the final result:

\begin{equation}
    \frac{1}{2}\Delta w^{T}H\Delta w = \frac{1}{2}w_{p}^{T}(H_{p,p}-H_{p,l}H_{l,l}^{-1}H_{l,p})w_{p}
\end{equation}

\subsection{Proving Equation \ref{eq:Hutchinson_Ex}}
\label{cha:Hutchinson_Ex}
Suppose we have a random vector $v$ sampled i.i.d. from a Rademacher distribution, then we have the following: \\

\begin{equation}
    Tr(H) = Tr(HI) = Tr(H\mathbf{E}[vv^{T}]) = \mathbf{E}[Tr(Hvv^{T})] = \mathbf{E}[v^{T}Hv]
\end{equation} \\

where $I$ refers to the identity matrix. Using the Hutchinson's Method, the relative Hessian trace can be estimated by taking multiple samples of $v^{T}Hv$ and computing its expectation.

\subsection{Experimental Settings and File Sizes}
\label{cha:Experimental_Settings}
We adopt the following settings for our experiments as shown in \autoref{Hyperparameters1}, \autoref{Hyperparameters2} and \autoref{Hyperparameters3}. We also present average file sizes, compression ratios and FLOPs in \autoref{File_Sizes_1} and \autoref{File_Sizes_2}.\\

}
\end{minipage}

\begin{table*}[]
\centering
\caption{Experimental Setup for Training}
\label{Hyperparameters1}
\begin{tabular}[t]{r|rrr}
\toprule
& \textbf{ResNet-32} & \textbf{ResNet-56} & \textbf{WRN-28-8} \\
\midrule
Epochs & 300 & 200 & 200 \\
Batch Size & 128 & 128 & 128 \\
Momentum & 0.9 & 0.9 & 0.875 \\
Weight Decay & 0.001 & 0.001 & 5e-4 \\
Initial LR & 0.05 & 0.05 & 0.1 \\
LR Scheduler & Cosine & Cosine & Cosine \\
\bottomrule
\end{tabular}
\end{table*}

\begin{table*}[]
\centering
\caption{Experimental Setup for Pruning and Fine Tuning}
\label{Hyperparameters2}
\begin{tabular}[t]{r|rrr}
\toprule
& \textbf{ResNet-32} & \textbf{ResNet-56} & \textbf{WRN-28-8} \\
\midrule
Epochs & 300 & 200 & 200 \\
Batch Size & 128 & 128 & 128 \\
Momentum & 0.9 & 0.9 & 0.9 \\
Weight Decay & 0.001 & 0.001 & 5e-4 \\
Initial LR & 0.01 & 0.01 & 0.0512 \\
LR Scheduler & Cosine & Cosine & Cosine \\
\midrule
Hessian Batch Size & 512 & 512 & 256 \\
$n_v$ & 300 & 300 & 300 \\
Prune Ratio (CIFAR10) & $\sim$0.77214 & $\sim$0.80870 & $\sim$0.76044 \\
Prune Ratio (CIFAR100) & $\sim$0.79354 & $\sim$0.83348 & $\sim$0.78081 \\
Prune Ratio Limit & 0.95 & 0.95 & 0.95 \\
\bottomrule
\end{tabular}
\end{table*} 

\begin{table*}[]
\centering
\caption{Experimental Setup for Quantizing}
\label{Hyperparameters3}
\begin{tabular}[t]{r|rrr}
\toprule
& \textbf{ResNet-32} & \textbf{ResNet-56} & \textbf{WRN-28-8} \\
\midrule
Epochs & 30 & 20 & 20 \\
Batch Size & 128 & 128 & 128 \\
Momentum & 0.9 & 0.9 & 0.875 \\
Weight Decay & 1e-4 & 1e-4 & 1e-4 \\
Initial LR & 1e-4 & 1e-4 & 1e-4 \\
LR Scheduler & Cosine & Cosine & Cosine \\
\midrule
Scheme & Uniform8 & Uniform8 & Uniform8 \\
Act. Percentile & 0 & 0 & 0 \\
Act. Range Momentum & 0.99 & 0.99 & 0.99 \\
Weight Percentile & 0 & 0 & 0 \\
\bottomrule
\end{tabular}
\end{table*}

\begin{table*}[]
\centering
\caption{CIFAR-10 Average Model File Sizes, FLOPs and Compression Ratios}
\label{File_Sizes_1}
\begin{tabular}[t]{r|rrr}
\toprule
\textbf{CIFAR-10} & \textbf{ResNet-32} & \textbf{ResNet-56} & \textbf{WRN-28-8} \\
\midrule
Original Size & 3761 KB & 6887 KB & 182590 KB \\
Pruned Size & 503 KB & 905 KB & 18127 KB \\
Quantized Size & 239 KB & 420 KB & 9276 KB \\
Original FLOPs & 0.14 G & 0.25 G & 6.73 G \\
Pruned FLOPs & 0.03 G & 0.03 G & 0.76 G \\
Compression Ratio & 10.26\% & 10.02\% & 9.88\% \\
\bottomrule
\end{tabular}
\end{table*}

\begin{table*}[]
\centering
\caption{CIFAR-100 Average Model File Sizes, FLOPs and Compression Ratios}
\label{File_Sizes_2}
\begin{tabular}[t]{r|rrr}
\toprule
\textbf{CIFAR-100} & \textbf{ResNet-32} & \textbf{ResNet-56} & \textbf{WRN-28-8} \\
\midrule
Original Size & 3807 KB & 6933 KB & 182956 KB \\
Pruned Size & 506 KB & 917 KB & 18442 KB \\
Quantized Size & 236 KB & 426 KB & 9169 KB \\
Original FLOPs & 0.14 G & 0.25 G & 6.73 G \\
Pruned FLOPs & 0.02 G & 0.03 G & 0.50 G \\
Compression Ratio & 9.95\% & 10.10\% & 10.01\% \\
\bottomrule
\end{tabular}
\end{table*}

\end{document}